\documentclass[10pt,twocolumn,letterpaper]{article}

\usepackage{algorithm}
\usepackage{algpseudocode}
\usepackage{graphicx}
\usepackage{amsmath}
\usepackage{amssymb}
\usepackage{multicol}
\usepackage{float}
\usepackage{array}
\usepackage{booktabs}
\usepackage{framed}
\usepackage{mdframed}
\usepackage{tikz} 
\usepackage{amsmath}
\usepackage{multicol,multirow}
\usepackage{setspace}
\usepackage{makecell}
\usepackage{xcolor}
\usepackage{cvpr}              

\definecolor{cvprblue}{rgb}{0.21,0.49,0.74}
\usepackage[pagebackref,breaklinks,colorlinks,allcolors=cvprblue]{hyperref}

\def\paperID{10144} 
\def\confName{CVPR}
\def\confYear{2025}

\title{~\includegraphics[height=20pt]{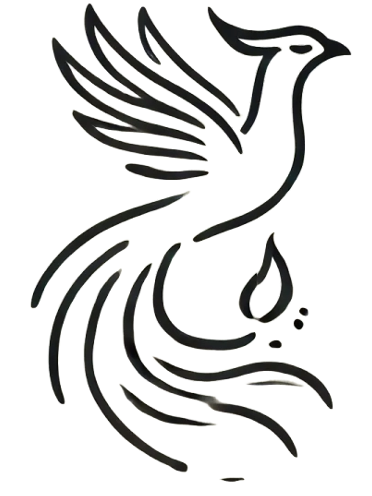}Phoenix: A Motion-based Self-Reflection Framework for 

Fine-grained Robotic Action Correction}
\author{
\textbf{Wenke Xia}\textsuperscript{1,2,\thanks{Work is done during internship at Shanghai AI Laboratory}},
\textbf{Ruoxuan Feng}\textsuperscript{1},
\textbf{Dong Wang}\textsuperscript{2},
\textbf{Di Hu}\textsuperscript{1,\thanks{Corresponding author}}
\vspace{0.5em}
\\
\textsuperscript{1}Gaoling School of Artificial Intelligence, Renmin University of China, Beijing\\
\textsuperscript{2} Shanghai AI Laboratory 
}

\begin{document}

\maketitle
\begin{abstract}

Building a generalizable self-correction system is crucial for robots to recover from failures.
Despite advancements in Multimodal Large Language Models (MLLMs) that empower robots with semantic reflection ability for failure, translating semantic reflection into ``\textbf{how to correct}'' fine-grained robotic actions remains a significant challenge.
To address this gap, we build the Phoenix framework, which leverages \textbf{motion instruction} as a bridge to connect high-level semantic reflection with low-level robotic action correction. 
In this motion-based self-reflection framework,
we start with a dual-process motion adjustment mechanism with MLLMs to translate the semantic reflection into coarse-grained motion instruction adjustment. To leverage this motion instruction for guiding ``\textbf{how to correct}'' fine-grained robotic actions, a multi-task motion-conditioned diffusion policy is proposed to integrate visual observations for high-frequency robotic action correction.
By combining these two models, we could shift the demand for generalization capability from the low-level manipulation policy to the MLLMs-driven motion adjustment model and facilitate precise, fine-grained robotic action correction.
Utilizing this framework, we further develop a lifelong learning method to automatically improve the model's capability from interactions with dynamic environments.
The experiments conducted in both the RoboMimic simulation and real-world scenarios prove the superior generalization and robustness of our framework across a variety of manipulation tasks. Our code is released at \href{https://github.com/GeWu-Lab/Motion-based-Self-Reflection-Framework}{https://github.com/GeWu-Lab/Motion-based-Self-Reflection-Framework}.

\vspace{-1em}

\end{abstract}
\section{Introduction}
\label{sec:intro}

\begin{figure}[t]
    \centering
		\includegraphics[width=0.9\linewidth]{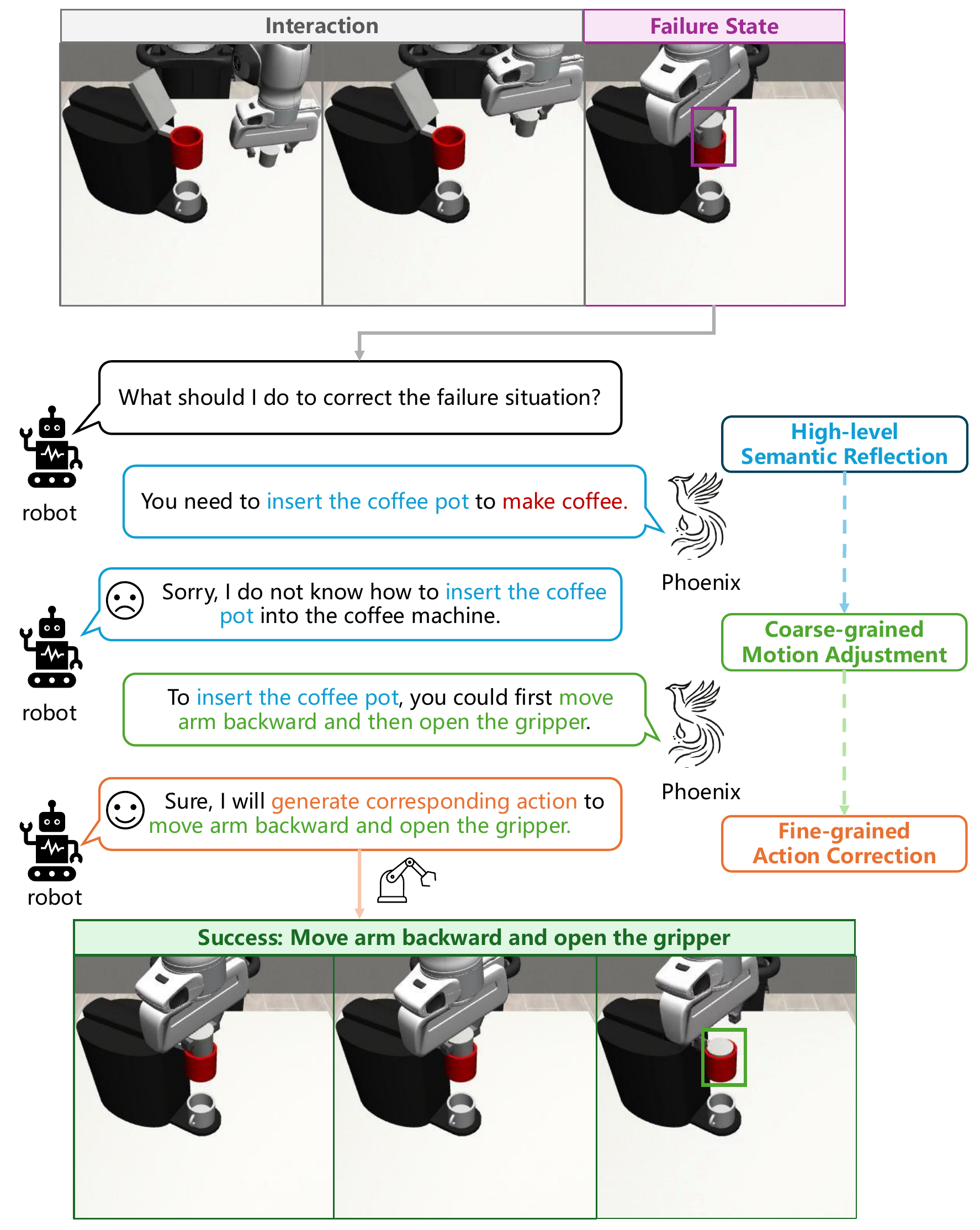}

    \caption{Our motion-based self-reflection framework utilizes coarse-grained motion instruction as a bridge to convert the high-level semantic reflection into fine-grained robotic action correction, thereby facilitating generalizable and precise action correction with perceptual and inferential capabilities of MLLMs.}
    \vspace{-2em}
    \label{fig:teaser}
\end{figure}

\begin{mdframed}[linewidth=1pt, linecolor=blue, backgroundcolor=gray!20]
``Failure is simply the opportunity to begin again, this time more intelligently.'' \hfill \textit{— Henry Ford}  
\end{mdframed}


Humans are naturally equipped with the ability to correct their behaviors by intentionally reflecting on actions that lead to failure.~\cite{lee2012neural,schank1995we}. By analyzing failure situations from high-level semantic reflection and low-level action correction perspective, humans can efficiently adapt to dynamic environments~\cite{flavell1979metacognition}.
To emulate the correction capability and foster a continuous cycle of self-improvement in robots, researchers~\cite{yu2020meta,schulman2017proximal,wang2023describe} have sought to develop self-reflection systems that enable robots to recover from and learn through their failure interactions.


Among them, some existing self-correction systems~\cite{schulman2017proximal,yokoyama2023asc,lu2024koi} leverage reinforcement learning to guide robots in correct low-level action execution through reward functions. However, the reliance on reinforcement learning limits the ability of these self-correction systems to generalize across long-horizon manipulation tasks, due to unstable training processes~\cite{horgan2018distributed} and the need for task-specific prior knowledge~\cite{yu2020meta,gu2023maniskill2}.
To construct a generalizable and stable self-correction system, recent works~\cite{wang2023describe,liu2024coherent,xia2024icra} borrow the inferential capability of Multi-modal Large Language Models (MLLMs) to propose closed-loop high-level semantic reflection framework for failure correction. Although these semantic self-reflection frameworks can decompose the failure correction process into semantic subgoals, they primarily rely on a predefined skill library to execute the detailed subgoals, which fails to utilize the generalization ability of MLLMs in fine-grained robotic action correction.


To maximize the generalization potential of MLLMs for action correction, we propose motion instruction as a bridge to convert high-level semantic reflection to fine-grained robotic action correction.
Motion instruction refers to coarse-grained robotic movement commands such as ``move arm backward'' and ``adjust gripper position''. Serving as an intermediate layer, motion instruction could provide general, low-frequency decision information for high-frequency robotic action execution, which makes it an excellent medium for embedding the knowledge of MLLMs into fine-grained action correction.
As shown in Figure~\ref{fig:teaser}, 
we decompose the semantic reflection knowledge into coarse-grained motion instruction adjustment to indicate ``how to correct'' fine-grained action for low-level policy execution.
This transition shifts the perceptual and decision-making requirements from low-level robotic policy to the MLLMs-driven motion adjustment model, thereby enabling generalizable, fine-grained robotic action correction.

Hence, in this work, we build the \textit{Phoenix} framework, a motion-based self-reflection framework designed to convert the semantic reflection of MLLMs into fine-grained robotic action correction. Initially, we develop a dual-process motion adjustment mechanism to ensure efficient prediction through a motion prediction module, while addressing failure with a motion correction module.
Concretely, we first utilize expert demonstration trajectories to train the motion prediction module for efficient motion instruction generation.
Despite its efficiency in generating initial instructions, this module often struggles to handle failure scenarios. To recover from failures, we collect a comprehensive failure correction dataset and fine-tune the motion correction module, which thoroughly provides adjusted motion instructions through a chain-of-thought approach. By integrating these two modules, the dual-process motion adjustment mechanism guarantees both robustness and efficiency, facilitating the generation of accurate motion instructions.
As the coarse-grained motion instructions only provide general and low-frequency guidance for robotic manipulation, we further design a multi-task motion-conditioned diffusion policy that integrates visual observations to translate motion instruction into precise, high-frequency action corrections for manipulation tasks.
Moreover, by leveraging these correction trajectories, we propose a lifelong learning method that iteratively enhances the model's capabilities through interaction, ensuring continuous improvements in performance and adaptability to dynamic environments.

To validate the efficacy of our framework, we conduct experiments across 9 contact-rich robotic manipulation tasks within the RoboMimic simulation~\cite{robomimic2021}. The results demonstrate that our method could provide more precise action 
 correction from failures through self-reflection and facilitate self-improvement through interactions with environments. 
 Further, we conduct two novel manipulation tasks with color disruption and position distribution disruption, proving the generalization ability of our framework.
 The real-world experiments also demonstrate the applicability and robustness of our approach in practical scenarios. 

\begin{figure*}[t]
    \centering
    \includegraphics[width=\textwidth]{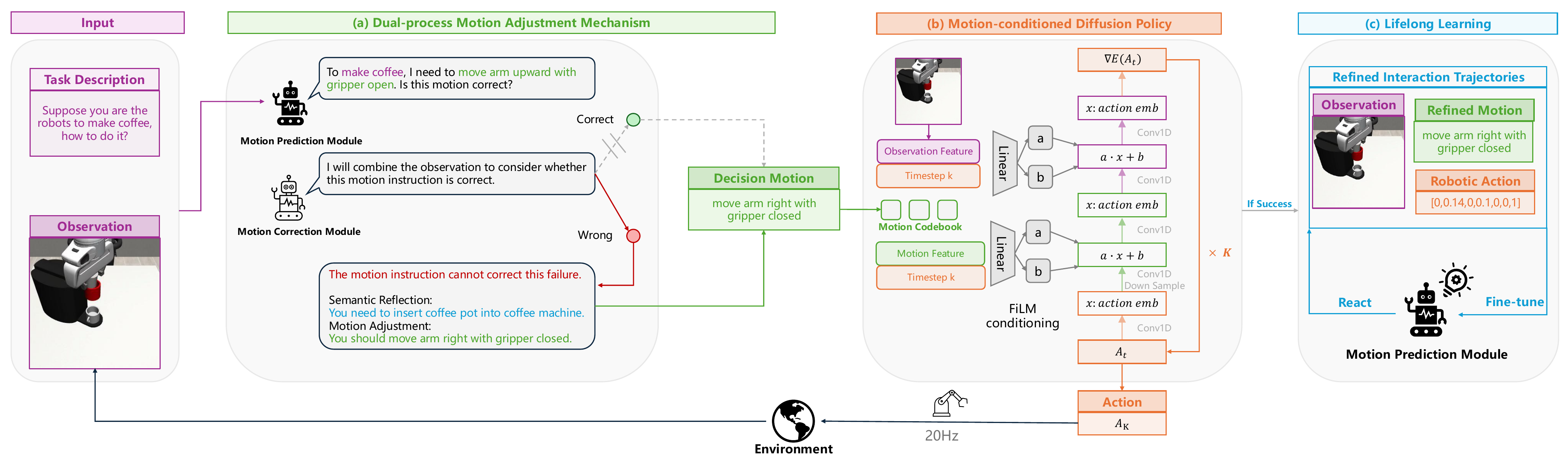}
    
    \caption{The pipeline of our motion-based self-reflection framework. (a) demonstrates the dual-process motion refinement mechanism which  leverages the motion prediction module for efficient motion instruction prediction and motion correction module for comprehensive failure correction. (b) illustrates the motion-conditioned diffusion policy which converts the low-frequency motion instruction guidance into high-frequency robotic action. The lifelong learning approach in (c) iteratively enhance the ability of the motion prediction module from the refined interaction trajectories.}
    
    \label{fig:pipeline}
    \vspace{-2em}
\end{figure*}

\section{Related Work}
\subsection{Robotic Self-correction Systems}
Self-correction serves as a crucial mechanism enabling robots to recover from failures. To achieve self-correction on the low-level robotic action, Reinforcement Learning~\cite{pateria2021hierarchical,barto2003recent} is proposed to guide robots in adjusting behaviors via reward signals. However, reinforcement learning strategies encounter difficulties in intricate robotic environments primarily due to learning inefficiency. Borrowing the common knowledge of MLLMs, Raman~et~al.~\cite{raman2022planning} proposes to build a semantic self-reflection system for long-horizon task planning. To facilitate interaction with environments, some efforts generate robotic action through simulation APIs~\cite{wang2023describe,wang2023voyager,liu2024coherent} and predefined action skill libraries~\cite{liang2023code,liu2024self}. 
However, the reliance on predefined low-level skill libraries makes semantic self-reflection frameworks fail to directly provide fine-grained action correction. To provide low-level action feedback for robotic manipulation, recent works~\cite{xiong2024aic,liu2024self,dai2024racerrichlanguageguidedfailure} suggest adjusting end-effector poses to refine actions. Nonetheless, these techniques are primarily restricted to simple manipulation tasks that employ motion planning and fail to generalize to contact-rich manipulation scenarios. In this work, we utilize motion instruction as an intermediate layer to guide robotic action correction, borrowing the perceptual and inferential capability of MLLMs for fine-grained robotic action correction.

\subsection{Robotic Manipulation Policy}
The development of generalizable strategies for robotic manipulation remains a persistent challenge in robotics research. ACT~\cite{zhao2023learning} is proposed to predict action sequences to ensure temporal alignment of actions, while the Diffusion Policy~\cite{chi2023diffusion} addresses multi-modal action distributions to enhance robust manipulation. Driven by the development of foundation models~\cite{touvron2023llama,li2024llava}, recent works~\cite{driess2023palm,brohan2023rt,liang2023code,zhang2024sam,pang2024depth,zeng2024learning} leverage the world knowledge of MLLMs to facilitate task decomposition and planning for robotic manipulation. Concurrently, other works~\cite{brohan2022rt,padalkar2023open,team2024octo,kim2024openvla} collect large-scale robotic manipulation demonstrations to train generalizable language-conditioned manipulation policy across different robots and tasks. Moreover, RT-H~\cite{belkhale2024rt} employs detailed motion commands instead of semantic language inputs, fostering flexible and generalizable manipulation capabilities through the advanced perceptual abilities of MLLMs.
However, the prohibitive costs associated with robotic data collection limit the scalability of existing imitation learning approaches.
To address this, we propose the motion-based self-reflection framework to enable autonomous self-improvement through continuous interaction with environments without human intervention.

\section{Motion-based Self-Reflection Framework}

\subsection{Challenges in Robotic Self-Corrction Model}

Building a generalizable and robust self-correction system is a key component in achieving failure correction for robots. Multi-modal Large Language Models (MLLMs) have already been applied to the construction of robotic self-reflection framework to recover from failures. However, existing systems mainly focus on semantic reflection, and their application to fine-grained action correction still faces the following two issues:
\begin{itemize}
    \item How to enable MLLMs to understand manipulation tasks and provide detailed correction information?
    \item How to convert the correction information provided by MLLMs into precise, high-frequency robotic actions?
\end{itemize}

To address these issues, we propose the \textit{Phoenix} framework, a motion-based self-reflection framework that integrates a dual-process motion adjustment mechanism and a multi-task motion-conditioned diffusion policy as illustrated in Figure~\ref{fig:pipeline}.
As detailed in Section~\ref{section:hierarchical}, the dual-process motion adjustment mechanism is developed to ensure efficient and accurate motion instruction generation. Further, the motion-conditioned diffusion policy is proposed to translate the coarse-grained motion instructions into precise robotic actions, as explained in Section~\ref{section:policy}.
Based on the refined manipulation trajectories, we propose a lifelong learning approach to facilitate robotic self-improvement, as outlined in Section~\ref{section:self_improvement}.


\subsection{Dual-process Motion Adjustment Mechanism}\label{section:hierarchical}


The dual-process motion adjustment mechanism is designed to ensure efficient motion prediction through a motion prediction module, while comprehensively addressing failure with a motion correction module.
Given the observation $O$ and task description $T$, we first train a Motion Prediction Module (MPM) with expert demonstration dataset $D_e$ to generate initial motion instruction $m_i$. However, the MPM trained on expert demonstrations struggles to handle failure situations. Thus, we construct a comprehensive failure correction dataset $D_c$ to fine-tune the Motion Correction Module (MCM), enabling it to analyze the failure situation and adjust $m_i$ with a chain-of-thought approach.
If $m_i$ is deemed correct, we adopt it as the decision motion instruction $m_d$ for further robotic action prediction. Otherwise, we employ the MCM to analyze the failure situation and generate adjusted motion instruction $m_a$ as decision motion instruction $m_d$. Through the guidance of $m_d$, our motion-based diffusion policy can generate high-frequency corrections to the robotic actions.
As described in Algorithm~\ref{alg:motion_correction}, we establish the dual-process motion adjustment mechanism to guarantee the efficiency and accuracy of motion instruction generation for fine-grained robotic action prediction.



\begin{algorithm}
\caption{Self-Reflection w/. Dual-Process Motion Adj.}
\label{alg:motion_correction}
\begin{algorithmic}[1]
\Require Task description $T$, Observation $O$, Environment $E$, Motion prediction module $MPM$, Motion correction module $MCM$, Motion-conditioned diffusion policy $\pi$, Exploration timestep $K$,

\State $O_1 \gets E.reset()$
\For{$k = 1$ \textbf{to} $K$}    
    \State $m_i \gets MPM(O_k,T)$
    \State $failure\_flag, semanic\_info \gets MCM(O_k,m_i)$
    \If{$failure\_flag$ is \textbf{true}}
        \State $m_a \gets MCM(O_k,sematic\_info)$
        \State$m_d \gets m_a$
    \Else
        \State $m_d \gets m_i$
    \EndIf
    \State $a \gets \pi(O,m_d)$
    \State $O_{k+1} \gets E.step(a)$
\EndFor

\end{algorithmic}
\end{algorithm}

\textbf{Motion Prediction Module (MPM).} 
To fully harness the perceptual and decision-making capabilities of MLLMs for efficient motion instruction prediction, we develop a motion instruction dataset from the expert demonstrations dataset $D_e$ to fine-tune MLLMs for robotic manipulation tasks.
To construct the expert dataset, we filter the robotic action to get dominant motion from expert demonstration with a threshold, generating a set of motion instructions that include arm direction and gripper control. In practice, we find that separating arm direction instruction and gripper control instruction would cause the misalignment between textual motion instruction and fine-grained robotic action. To address this issue, we combine the direction movement with gripper control, resulting in unified instruction formats for motion instruction such as ``move arm right with gripper closed''.
In addition, we add the ``make slight adjustments to gripper position'' instruction to model the minor robotic actions below the threshold. Through the automatic construction method, we build 37 types of motion instructions as guidance for further robotic action prediction. By training on the expert dataset, the MPM acquires an understanding of robotic manipulation tasks and could efficiently generate an initial motion instruction $m_i$.



\textbf{Motion Correction Module (MCM).}
During interactions with the environment, robots may execute incorrect actions, leading to a failure situation in the task. However, the MPM trained on success expert data often struggles to recover from these failure scenarios. 
Thus, we develop the motion correction module, to identify failure scenarios and correct behaviors from such situations. As shown in Figure~\ref{fig:pipeline}(a), the MCM would evaluate the initial motion instruction $m_i$ and conduct a dual process based on the evaluation results to achieve efficient and accurate motion instruction adjustment. Once encountering the failure situation, the MCM would first analyze the type of failure and derive a semantic-level correction goal, such as ``insert the coffee pot into the coffee machine''. Based on this correction goal, the MCM further adjusts the motion instructions with its learned failure-correction knowledge, ultimately generating an accurate motion instruction through a hierarchical chain-of-thought approach.



To equip MCM with the capabilities for failure detection and correction, we construct a comprehensive correction dataset as illustrated in Figure~\ref{fig:data}. This dataset includes three types of feedback data, encompassing both semantic and motion perspectives, and is categorized as follows:

\begin{figure}[t]
    \centering
		\includegraphics[width=0.9\linewidth]{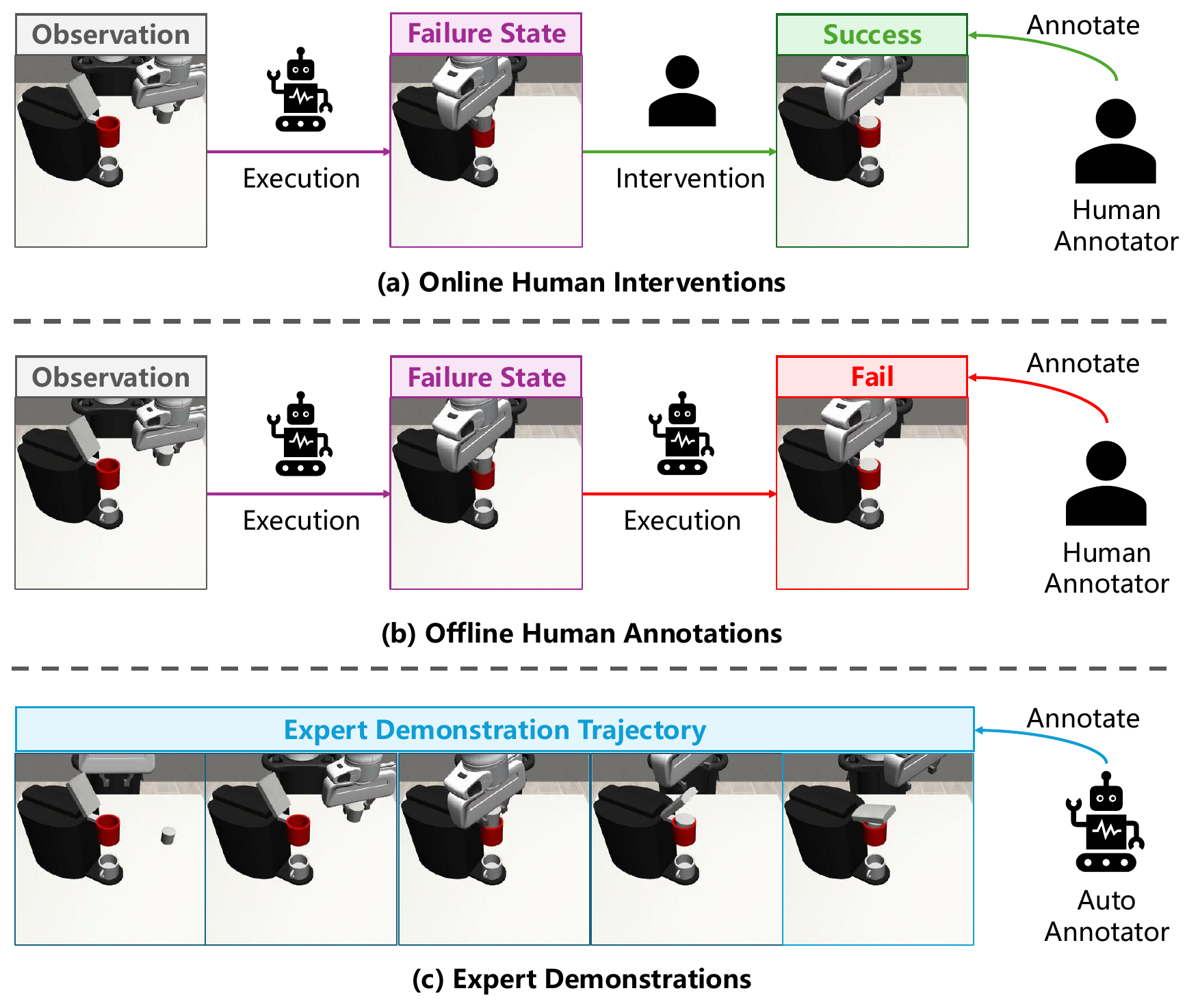}

    \caption{Illustrations of our correction data: online human interventions, offline human annotations, and expert demonstrations.}
    \label{fig:data}
    \vspace{-2em}
\end{figure}

\begin{itemize}
    \item \textbf{Online Human Intervention.} We implement the human-in-the-loop method for trajectory collection. We first deploy the motion prediction model to interact with environments, then manually intervene to correct the motion instructions whenever the agent encounters failure situations.
    This method could collect accurate and high-quality motion correction data to ensure task completion. However, it requires frequent manual interactions with the environment, which leads to significant time consumption and makes it difficult to collect large-scale data.
    \item \textbf{Offline Human Annotation.} 
    We utilize the motion prediction model to gather trajectory data, periodically sampling trajectories and annotating them with semantic reflections and motion correction details. While the accuracy of offline annotated data cannot be guaranteed due to its inability to interact with environments for verification, this method offers a significant volume of annotations.
    \item \textbf{Expert Demonstration.} We automate annotations on expert trajectories. Since these trajectories are successful, this data is used to provide accurate motion information to enhance the model's motion prediction capabilities.
\end{itemize}





By fine-tuning MCM on this dataset, we enhance the MCM to thoroughly comprehend various types of failure situations and provide motion instruction corrections. Through the integration of MPM and MCM, the dual-process motion adjustment mechanism enables efficient motion instruction generation while ensuring comprehensive correction in failure situations.






\subsection{Motion-conditioned Diffusion Policy}\label{section:policy}
As the motion instruction only provides general and low-frequency guidance for manipulation, we train a multi-task motion-conditioned diffusion policy $\pi$ to convert the motion instructions into precise, high-frequency robotic actions. This policy takes observations $O$ and decision motion instructions $m_d$ to output robotic actions $a$. To ensure the policy adheres to the motion instruction, we make adjustments as depicted in Figure~\ref{fig:pipeline}(b):


First, we observe that existing pre-trained language models often struggle to capture the discriminative features of various motion instructions. This limitation hampers their ability to follow various motion instructions.
To address this issue, we introduce a learnable motion codebook designed to provide discriminative features for motion instructions. For a given decision motion instruction $m_d$, the codebook would retrieve the corresponding motion feature to facilitate accurate robotic action prediction. 

Further, we find that the direct concatenation of observation representation and motion instruction feature would cause the diffusion policy to prefer to rely on the vision information for action prediction, thereby hindering the effectiveness of the motion instruction guidance. To address this issue, we take the observation representation and motion instruction feature as separate conditions in different stages of the diffusion policy, allowing the model to better learn the guidance information from the motion instruction and thereby promote precise action correction.

By integrating these two adjustments, we train the diffusion policy for action prediction with the following loss:
\begin{equation}
\mathcal{L} = \text{MSE}(\mathcal{E}^k, \pi(\mathcal{O}, \mathcal{M}, \mathcal{A}^0 + \mathcal{E}^k, k)), \label{eq:diffusion}
\vspace{-1em}
\end{equation}
where $\mathcal{O}$ is the observation representation, $\mathcal{M}$ is the motion instruction feature, $\mathcal{A}^0$ is the ground truth robotic action, $\mathcal{E}^k$ indicates the random noise at the denoising iteration $k$.
Through minimizing the loss function in Eq~\ref{eq:diffusion}, the diffusion policy $\pi$ could effectively predict precise, high-frequency robotic action guided by motion instruction.




\subsection{Action Correction for Lifelong Learning }\label{section:self_improvement}
The dual-process motion adjustment mechanism leverages the MPM to efficiently predict motion instructions and the MCM to adjust them with a comprehensive chain-of-thought approach. However, the reliance on the chain-of-thought poses challenges in adapting to real-time scenarios due to its time-consuming.  Furthermore, the collection of manipulation data and correction data is exceedingly labor-intensive.
Thus, we propose a lifelong learning method that equips the MPM with both motion prediction and failure correction capabilities through learning from the refined interaction trajectories as illustrated in Figure~\ref{fig:pipeline}(c), which enhances our model to adapt and react quickly to the environment without human intervention. 

Benefiting from the motion-conditioned diffusion policy which could adhere to the motion instruction to generate task-aware robotic action, we can enhance the robot's capabilities through only improving the MPM informed by the refined interaction trajectory. To address the issue of catastrophic forgetting, we mix the refined interaction trajectory with expert demonstration for co-fine-tuning, allowing the model to simultaneously learn failure correction and enhance the motion prediction capabilities.
Through updates from refined interaction trajectories, our model can achieve self-improvement by learning from the knowledge of the motion correction module, achieving fast and accurate manipulation for contact-rich manipulation tasks.

\begin{table*}[t]
\centering
\resizebox{\textwidth}{!}{
\begin{tabular}{ccccccccccc}
\toprule
Methods & Coffee\_D0 & Coffee\_D1  & Stack\_D0 & Stack\_D1 & StackThree\_D0 & StackThree\_D1 & Threading\_D0 & ThreePieceAssembly\_D0 & ThreePieceAssembly\_D1 & Mean\\ \midrule

OpenVLA~\cite{kim2024openvla} & 42\% & 18\% & 84\% & \textbf{86\%} & 36\% & \textbf{20\%}  & 20\%  & 28\% & \textbf{8\%} & 38.0\%    \\

Task-conditioned & 66\% & 24\% & 88\% & 68\% & 30\% & 6\% & 74\% & 20\% & 0\% & 41.8\%    \\
Subgoal-conditioned & 76\% & 26\% & 88\% & 74\% & 24\% & 6\% & \textbf{78\%} & 20\% & 2\% & 43.8\%  \\
Motion-conditioned & 68\% & 32\% & 92\% & 84\% & 38\% & 16\% & 58\% & 30\% & 4\% & 46.9\%  \\
\midrule

Subgoal Self-reflection & 80\% & 32\% & 88\% & 78\% & 32\% & 6\% & 80\% & 34\% & 2\% & 48.0\%    \\

Phoenix (Ours)&   \textbf{94\%} & \textbf{48\%} & \textbf{96\%} &\textbf{86\%} & \textbf{50\%} & \textbf{20\%} & 68\% & \textbf{52\%} & 6\% & \textbf{57.8\%} \\

\midrule
Human Intervention (Oracle) & 100\% & 100\% & 100\% & 90\% & 70\% & 40\% & 100\% & 70\% & 40\% & 78.9\%    \\
\bottomrule
\end{tabular}}
    \caption{Comparison experiments results across 9 manipulation tasks in RoboMimic Simulation. The results demonstrate that our motion-based self-reflection method achieves better performance by facilitating precise correction of fine-grained robotic actions.}
    \label{tab:main_results}
    \vspace{-1em}
\end{table*}

\section{Experiments}

To comprehensively evaluate our framework, we propose experiments to answer the following questions:
\begin{itemize}
    \item  Does our motion-guided self-reflection model enhance the precision of action correction? Section~\ref{exp:performance}
    \item Can our model achieve lifelong learning from interaction with environments? Section~\ref{exp:self_improvement}
    \item Does our framework can generalize across novel tasks? Section~\ref{exp:generalization}
    \item Can our framework ensure reliability and robustness in real-world scenarios? Section~\ref{exp:real_world}
\end{itemize}

\subsection{Experiment Settings}
In this work, we conduct experiments on 9 contact-rich manipulation tasks in RoboMimic~\cite{robomimic2021}, ranging from long-horizon tasks like ThreePieceAssembly to fine-grained manipulation tasks like Threading.
To transform high-level semantic information into motion instructions, we filter expert demonstrations to obtain over 160,000 pairs of motion instructions and observations. The dataset includes 37 types of motion instructions, which are utilized to fine-tune the LLaVA-v1.5 model~\cite{liu2023llava} as the motion prediction module.
Furthermore, to develop the motion correction module that integrates semantic comprehension and motion instruction adjustment, we collect correction data comprising 3,644 online human intervention data, 7,365 offline human annotations, and 6,378 expert demonstrations. We filter out the correction dataset to balance the proportion of various failure situations to enhance the model's correction capabilities. 
Ultimately, to translate motion instructions into precise robotic actions, we train a multi-task motion-conditioned diffusion policy using a learnable motion instruction codebook, incorporating 500 demonstrations per task. During inference in simulation, our dual-process motion adjustment mechanism would provide motion instruction at 5\textit{Hz}, and the diffusion policy would extend the motion instruction with visual observations to a 20\textit{Hz} action sequence to control the robot. For each task, we conducted 50 trials and report the average success rate. More implementation details could refer to Supp.A.


\subsection{Performance of Motion Self-Reflection Model}\label{exp:performance}



\begin{table*}[t]
\centering
\resizebox{\textwidth}{!}{
\begin{tabular}{ccccccccccc}
\toprule
Methods & Coffee\_D0 & Coffee\_D1  & Stack\_D0 & Stack\_D1 & StackThree\_D0 & StackThree\_D1 & Threading\_D0 & ThreePieceAssembly\_D0 & ThreePieceAssembly\_D1 & Mean\\ \midrule
Motion-conditioned & 68\% & 32\% & 92\% & 84\% & 38\% & 16\% & 58\% & 30\% & 4\% & 46.9\%  \\

Expert-Correction Mixture & 74\% & 36\% & 94\% & 86\% & 38\% & 22\% & 64\% & 30\% & 2\% & 49.6\%    \\


\midrule

Expert-Correction Mixture with Self-Reflection & 76\% & 30\% & 92\% & \textbf{90\%} & 46\% & \textbf{26\%} & 64\% & 34\% & 4\% & 51.3\%  \\

Phoenix (Ours)&   \textbf{94\%} & \textbf{48\%} & \textbf{96\%} & 86\% & \textbf{50\%} & 20\% & \textbf{68\%} & \textbf{52\%} & \textbf{6\%} & \textbf{57.8\%} \\

\bottomrule
\end{tabular}}
    \caption{The ablation results of our dual-process motion adjustment mechanism. The results prove that our model, which separates motion prediction module and motion correction module, can provide more precise motion adjustment for action execution in manipulation tasks.}
    \label{tab:Ablation}
    \vspace{-1em}
\end{table*}

\begin{figure*}[t]
    \centering
    \includegraphics[width=\linewidth]{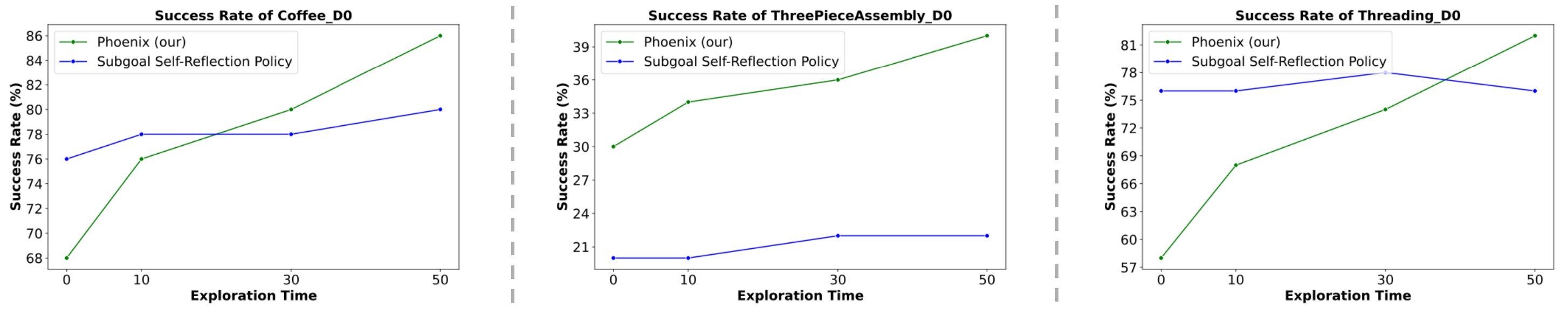}

    \caption{The lifelong learning results. The results prove that our motion-based self-reflection method could iteratively improve performance through interactions with environments.}

    \label{fig:continual}
    \vspace{-1em}
\end{figure*}

\subsubsection{Comparison Results}
To evaluate our motion-based self-reflection framework, we compare our framework with other approaches.
To ensure fairness, all our comparison methods are trained on the expert data from the simulation environment, with the decision model using LLaVA-v1.5 and the underlying policy employing a diffusion policy.

\begin{itemize}
    \item \textbf{OpenVLA~\cite{kim2024openvla}.} We fine-tune the OpenVLA model to provide baseline performance for multi-task experiments.

    \item \textbf{Task-conditioned policy.} We take the task description as the condition for diffusion policy without the reflection framework, as a variance of RT-1~\cite{brohan2022rt} and Octo~\cite{team2024octo}.
    \item \textbf{Subgoal-conditioned policy.} We fine-tune a LLaVA-v1.5 to predict subgoals at 5Hz, which are utilized as the condition for diffusion policy without reflection framework. This method borrows the semantic comprehension capabilities of MLLMs, and is implemented as a variance of PaLM-E~\cite{driess2023palm} with an individual diffusion policy.
    \item \textbf{Motion-conditioned policy.} We fine-tune a LLaVA-v1.5 as the motion prediction model to provide motion instructions at 5Hz, using these predictions to condition the diffusion policy without the reflection framework. This method employs the perceptual and inferential capacities of MLLMs, realized as a variation of RT-H~\cite{belkhale2024rt} with an individual diffusion policy.
    \item \textbf{Human Intervention.} We manually correct the wrong motion instructions for the motion-conditioned policy. This method provides an upper bound on the performance of self-reflection methods. Due to labor costs, the results are presented as average success rates across 10 trials.
    \item \textbf{Subgoal Self-reflection.} We fine-tune a LLaVA-v1.5 as subgoal self-reflection model and apply it to the subgoal-condition policy. This method is designed to validate the effectiveness of the semantic self-reflection model.
\end{itemize}

As shown in Table~\ref{tab:main_results}, we first compare three different condition methods. Borrowing the perception and inferential ability of MLLMs, the subgoal-conditioned and motion-conditioned policies are better than the task-conditioned policies. The results prove the potential applications of MLLMs in various complex robotic manipulation tasks.

Focusing on specific tasks, we observe that the motion-conditioned policy excels in long-horizon tasks such as StackThree\_D0 and ThreePieceAssembly\_D0. However, this policy depends on consistent and accurate motion instruction predictions, which poses challenges in fine-grained manipulation tasks like Threading\_D0. 

By providing correction subgoals, the subgoal self-reflection method consistently outperforms the subgoal-conditioned policy, particularly in long-horizon manipulation tasks such as ``StackThree\_D0'', which demonstrates the efficacy of the self-reflection framework.

The OpenVLA model demonstrates strong performance in certain long-horizon tasks, leveraging its end-to-end action token prediction capability. However, the lack of observation history and action chunking poses significant challenges in handling complex, fine-grained manipulation tasks like Threading\_D0.

Notably, our Phoenix method achieves more substantial improvements than the subgoal self-reflection method, demonstrating the effectiveness of motion-conditioned method in long-horizon sequential tasks and fine-grained manipulation tasks.
Benefiting from our motion-based correction method, agent could correct fine-grained action through motion instruction adjustment while the subgoal-conditioned self-reflection model fails to recover from most failure situations.
Besides, the human intervention method achieves high success rates across multiple tasks, demonstrating that our motion-conditioned diffusion policy can effectively adhere to motion instructions for manipulation tasks. This result indicates that our method can perform well under the correct motion instructions, showcasing the significant potential of motion-conditioned self-reflection.


\begin{figure*}[th]
    \centering
    \includegraphics[width=\linewidth]{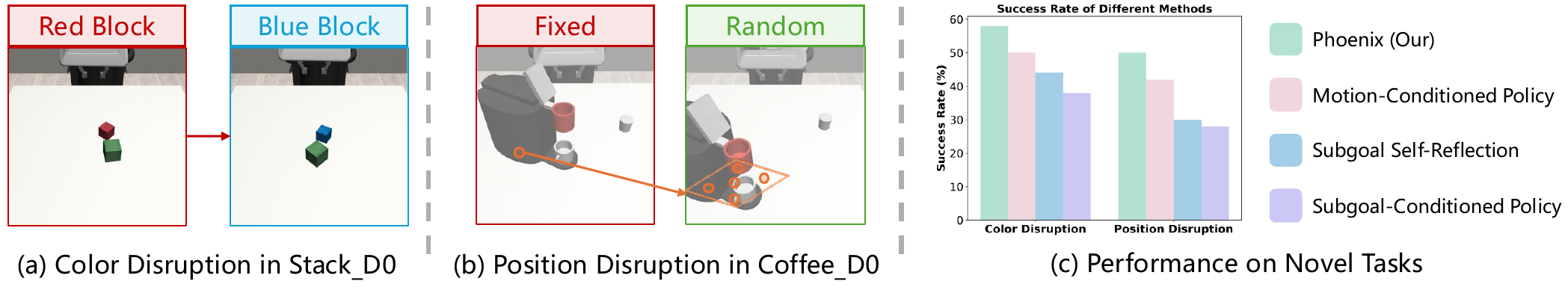}

    \caption{
    In the color disruption setting, we replace the \textcolor{red}{red block} with the \textcolor{blue}{blue block} in the Stack\_D0 task as shown in (a). In the position disruption setting, we change the position of the coffee machine from a fixed point \protect\tikz{\protect\filldraw[color=orange, fill=white] (0,0) circle (2pt);} to a random position from the rectangle \protect\tikz{\protect\draw[color=orange] (0,0) rectangle (0.15,0.15);} in the Coffee\_D0 task as illustrated in (b). The results in (c) prove that our framework could generalize well to these novel task settings.
    }

    \label{fig:disruption}
    \vspace{-1em}
\end{figure*}

\subsubsection{Ablation Results}


In this work, we propose a motion prediction module to provide initial motion instruction, and a motion correction module to provide fine-grained motion correction. Drawing upon prior research~\cite{liu2024regmix,ye2024data}, data mixture proportions could influence the efficacy of LLMs. 
In this section, we investigate whether integrating expert demonstrations with correction dataset, could also enhance the perception and decision-making capabilities of MLLMs for robotic manipulation with the following ablation methods: 
\begin{itemize}
    \item \textbf{Expert-Correction Mixture.} We mix the expert demonstration and correction data for co-training the motion prediction model.
    \item \textbf{Expert-Correction Mixture with Self-Reflection.} We mix the expert demonstration and correction data for co-training a unified model to provide initial motion instruction and adjust the instruction.
\end{itemize}

As illustrated in Table~\ref{tab:Ablation}, the results show that co-training with mixture data yields superior performance compared to models trained exclusively on expert demonstration data. This indicates that combining various types of feedback data can enhance the decision-making and perception capabilities of MLLMs. It also validates the viability of our approach to achieving self-improvement through interaction.

Besides, the mixture training model with self-reflection performs better than the one without self-reflection, which suggests that our designed motion-based self-reflection method can enhance the decision-making capabilities of robots and facilitate the correction of fine-grained actions. 

However, we find that utilizing the mixture of data to train a unified model to serve as both the motion prediction module and motion correction module fails to provide accurate correction information compared to our separated motion correction module. This suggests that the mixture training strategy may not fully leverage the strengths of each dataset to achieve better correction effects under the significant data scale discrepancies (160,000 expert demonstrations vs. 16,000 feedback data). The results indicate that our dual-process motion adjustment mechanism can effectively leverage the expert demonstration and correction dataset, leading to accurate motion instruction adjustment.

We also provide ablation results of our designed motion codebook in Supp.B.

\begin{figure*}
    \begin{minipage}[!t]{0.55\textwidth}
        \centering
        \includegraphics[width=\linewidth]{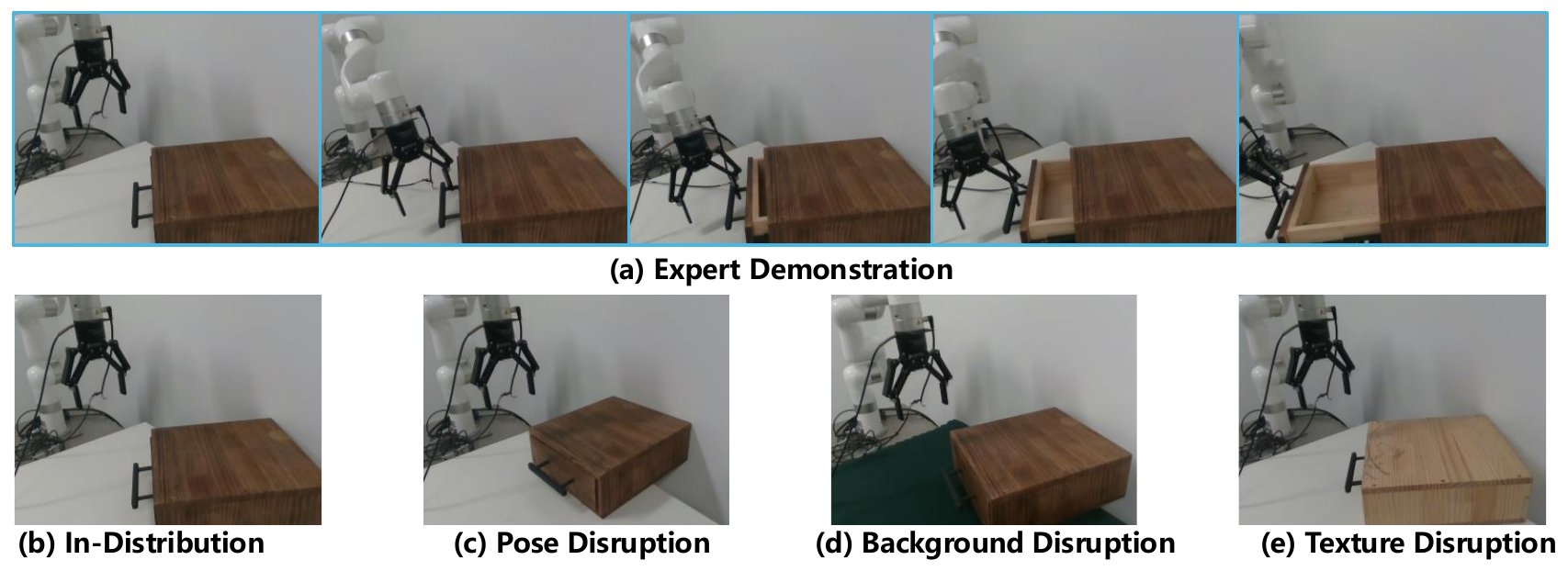}
         \vspace{-1em}
        \caption{The real-world experiments with different variations.}
        \label{fig:real}
    \end{minipage}
    \hspace{0.05\textwidth}
    \begin{minipage}[!t]{0.35\textwidth}
    \begin{center}

    \small
    \begin{tabular}{|c|c|c|c|c|}
      \hline

      Model & In-Dis. & Pose Dis. & Bg. & Tex. \\
      \hline
      OpenVLA & 55\% & 30\% & 35\% & 45\% \\
      Task & 60\% & 25\% & 25\% & 45\% \\
      Motion & 60\% & 35\% & 30\% & 50\% \\
      Ours & 75\% & 55\% & 45\% & 65\% \\
      \hline
    \end{tabular}
    \vspace{-1em}
    \captionof{table}{The real-world experiment results.}
    \label{tab:real_results}
    
    \begin{tabular}{|c|c|c|c|}

      \hline

      Task & Motion & 10 rollout & 30 rollout \\
      \hline
      In-Dis. & 60\% & 65\% & 75\% \\
      Pose Dis. & 35\% & 45\% & 50\% \\
      \hline
    \end{tabular}
     \vspace{-1em}
    \captionof{table}{The lifelong learning results}
    \label{tab:real_lifelong}
        
    \end{center}

    \end{minipage}
    \vspace{-1em}
\end{figure*}

\subsection{The Performance of Lifelong Learning}\label{exp:self_improvement}
In this section, we explore whether our Phoenix framework can facilitate lifelong learning through interactions.
Concretely, we deploy the motion self-reflection model to interact within the environments and utilize the successful trajectories to iteratively fine-tune our motion prediction model after 10, 30, and 50 rollouts.
To avoid catastrophic forgetting, we combine 20 expert demonstrations to co-fine-tune the motion prediction module. 

We compare the lifelong learning ability of our motion-based self-reflection model and subgoal-based self-reflection model. During testing, we record the average success rate over 50 trials. As shown in Figure~\ref{fig:continual}, the subgoal-based lifelong learning fails to enhance model performance during the exploration phase due to its inability to provide fine-grained action correction. In contrast, our method corrects underlying action execution during interactions, allowing the robot to better learn from the refined trajectories, thereby achieving self-improvement.

\subsection{Generalization to Novel Tasks}\label{exp:generalization}
In this section, we evaluate the generalization ability of our Phoenix framework in color disruption and position disruption novel tasks as shown in Figure~\ref{fig:disruption}. 
In the color disruption setting, we replace the red block with the blue block in the Stack\_D0 task to verify whether our model could generalize to object manipulation tasks with different visual characteristics.
In the position disruption setting, we change the fixed position of the coffee machine to a randomly placed position within a specific area in the Coffee\_D0 task to verify whether our method could generalize to unseen scenarios. 

For these novel tasks, although the subgoal-conditioned policy could predict correct high-level semantic subgoal for manipulation, this method fails to predict precise robotic action to complete the tasks. 
Due to its limitation of providing high-level semantic correction information, the subgoal self-reflection method fails to effectively leverage the knowledge of MLLMs for action correction to manipulation tasks.
In contrast, as shown in Figure~\ref{fig:disruption}(c), our motion-conditioned policy could generate fine-grained motion instruction to achieve generalizable manipulation benefiting from the perception and inferential capability of MLLMs. Besides, our method could achieve better performance on novel tasks by comprehensively refining the motion instruction with the motion-based self-reflection framework.





\subsection{Real-World Experiments}\label{exp:real_world}

In real-world scenarios, we conduct the challenging ``drawer open'' articulated object manipulation task as shown in Fig~\ref{fig:real}(a), where the robot needs to align gripper with handle through precise rotations to open the drawer. We utilize the spacemouse device to collect 100 expert demonstrations with 14 motion instructions (e.g.,``move arm right'',``rotate around x-axis''). We train a motion-conditioned diffusion policy to convert instructions into robotic actions. During the inference, we introduce human-in-the-loop interventions to manually correct failure situations to collect 20 corresponding refined interaction trajectories to train our motion correction module.   All models are only fine-tuned on the real-world data.

To validate generalization, we design 4 settings as shown in Fig~\ref{fig:real}(b-e). In the pose disruption setting, we change the pose distribution of the drawer. For the background disruption setting, the background color was modified to green. In the texture disruption setting, the texture of the drawer was altered to evaluate performance under significant visual variations.
The results in Tab~\ref{tab:real_results} demonstrate the generalization ability of our method. We also evaluate lifelong learning, the results in Tab~\ref{tab:real_lifelong} show that our model achieves self-improvement in real world. 

We also provide more real-world task experiments with a rule-based manipulation policy to prove the effectiveness of our motion-based self-reflection method in Supp.C.

\section{Conclusion}
In this work, we propose a motion-based self-reflection framework to convert the semantic reflection of MLLMs into fine-grained robotic action correction. Based on this framework, we further automatically improve the model's capability from interactions. 
We hope this motion-based self-reflection framework could bring insights for enhancing the generalization capabilities of agents in robotic manipulation tasks through the integration of MLLMs.


\section{Acknowledgement}

The project was supported by National Natural Science Foundation of China (NO.62106272), Shanghai AI Laboratory, National Key R\&D Program of China (2022ZD0160101) and  CCF-Zhipu.AI Large Model Innovation Fund.



{
    \small
    \bibliographystyle{ieeenat_fullname}
    \bibliography{main}
}

\end{document}


\maketitle

\section{Implementation Details}
\subsection{Dual-process Motion Adjustment Mechanism}

\textbf{Training Details.} 
In this mechanism, we construct a motion prediction module to efficiently obtain the initial motion prediction and a motion correction module to provide comprehensive motion adjustment.
All of our models are fine-tuned based on the LLaVA1.5 framework~\cite{liu2023llava}, which encompasses three essential components: (1) a vision encoder utilizing the capabilities of the CLIP-Large model~\cite{radford2021learning}, which operates at a resolution of 336x336 and utilizes a patch size of 14, (2) a two-layer MLP projector that facilitates the fusion of visual and linguistic modalities, and (3) a language model, derived from the open-source Vicuna-v1.5~\cite{vicuna2023}, building on the LLaMA2 foundation. We fine-tune the projector and train the LoRA layer~\cite{hu2021lora} across each transformer attention block. The learning rates are set at 1e-5 for the projector layer and 1e-4 for the LoRA layer, with a LoRA alpha of 256 and a dimension of 128. The motion prediction module undergoes training over five epochs on the motion instruction dataset with a batch size of 16. The motion correction module is trained for 20 epochs on the correction dataset, also with a batch size of 16.

\begin{figure}[t]
    \centering
    \includegraphics[width=\linewidth]{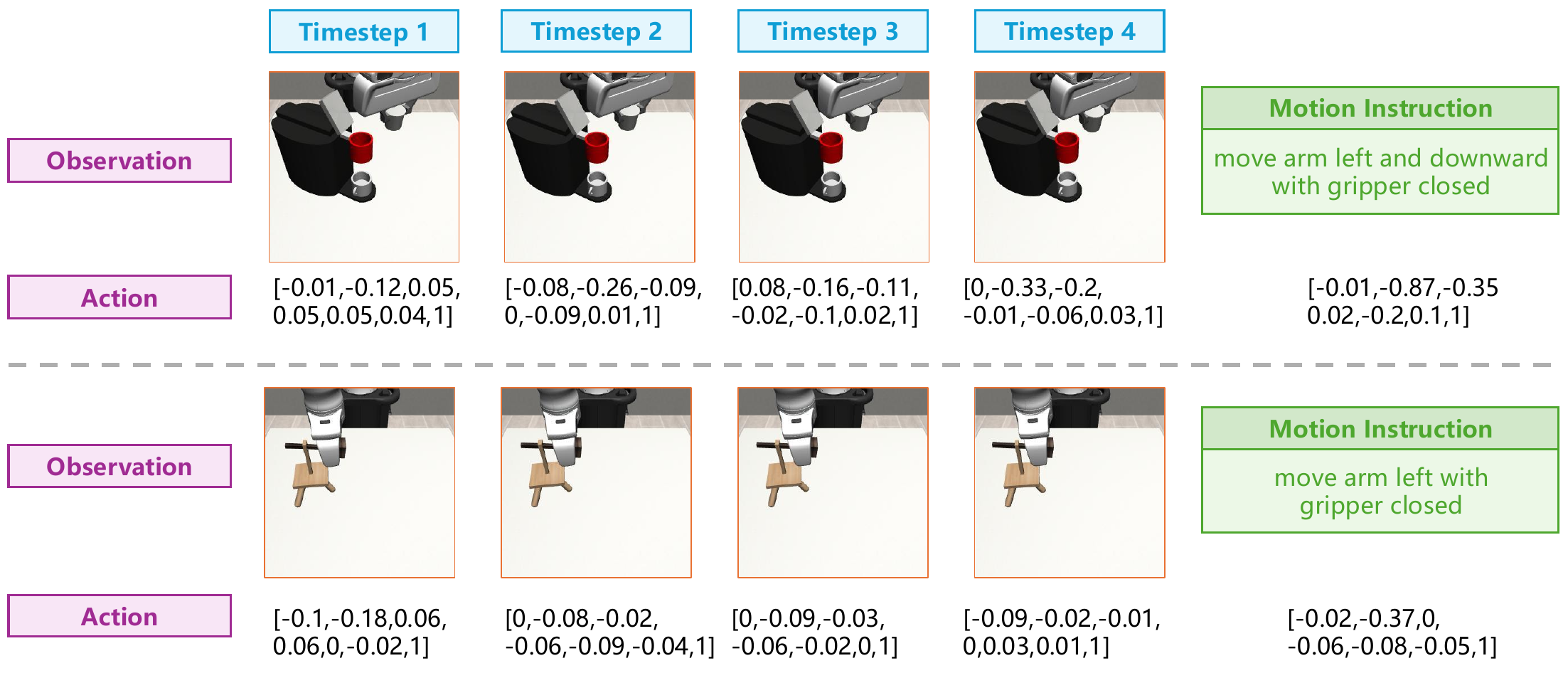}
    
    \caption{The motion instruction dataset annotation process. }
    \label{fig:annotation}

\end{figure}

\textbf{Motion Instruction Dataset from Expert Demonstrations.} To empower the motion prediction module with the comprehension of manipulation tasks, we construct a motion instruction dataset from expert demonstrations to obtain over 160,000 pairs of motion instructions and observations. Due to the limited inference speed of MLLMs, it is difficult to utilize the motion instruction from MLLMs for real-time robotic control. Therefore, during motion instruction annotations, we aggregate  4 timestep robotic actions to form a single temporal robotic action and annotate motion instruction. We demonstrate the automatic motion instruction annotation process in Figure~\ref{fig:annotation}. We first filter the temporal robotic action to obtain the dominant direction with a threshold of 0.3. If an action exhibits more than one direction exceeding the threshold, the top two directions are selected as the dominant directions. Further, we obtain the gripper action from the temporal robotic action and combine the gripper action with the motion instruction. Furthermore, we incorporate the instruction to "make slight adjustments to gripper position" to model the temporal robotic actions that fall below the threshold. Utilizing the automated construction methodology, we develop a diverse set of 37 distinct motion instructions. These instructions serve as a comprehensive guide for enhancing the precision of subsequent robotic action predictions.

\begin{figure}[t]
    \centering
    \includegraphics[width=\linewidth]{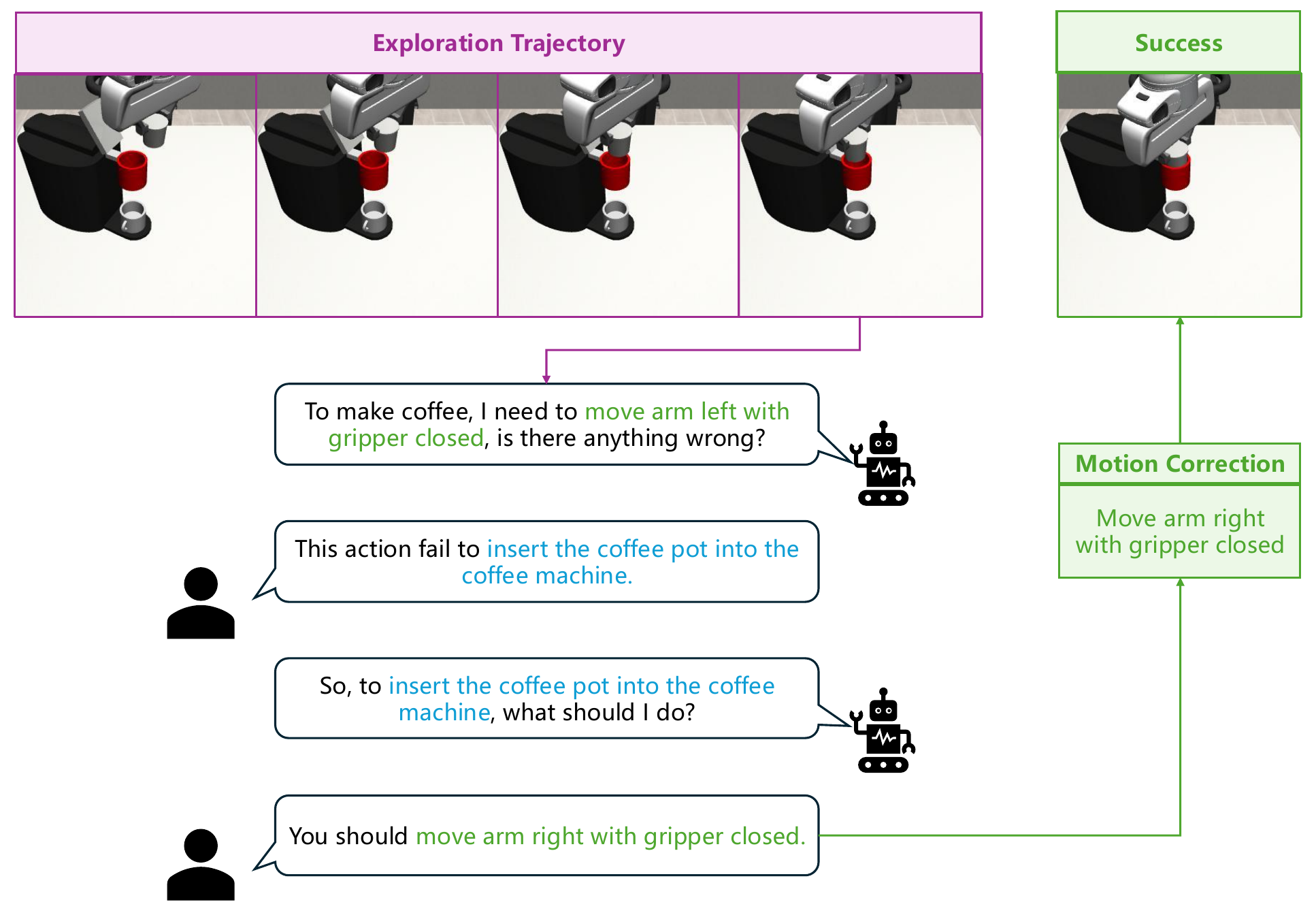}
    
    \caption{The demonstration of online human intervention data collection process. }
    \label{fig:online}

\end{figure}

\textbf{Motion Correction Dataset.}
To equip the motion correction module with the capabilities for failure correction, we build a comprehensive correction dataset to provide semantic reflection and motion adjustment annotation. 

The online human intervention data are collected as shown in Figure~\ref{fig:online}. When the robot interacts with the environment, the human checks the task execution situation in real-time. When a failure occurs, humans provide semantic reflection and adjust motion instructions. Consequently, the robot corrects its fine-grained actions based on the adjusted motion instruction through the implementation of a low-level diffusion policy.

\begin{figure}[t]
    \centering
    \includegraphics[width=\linewidth]{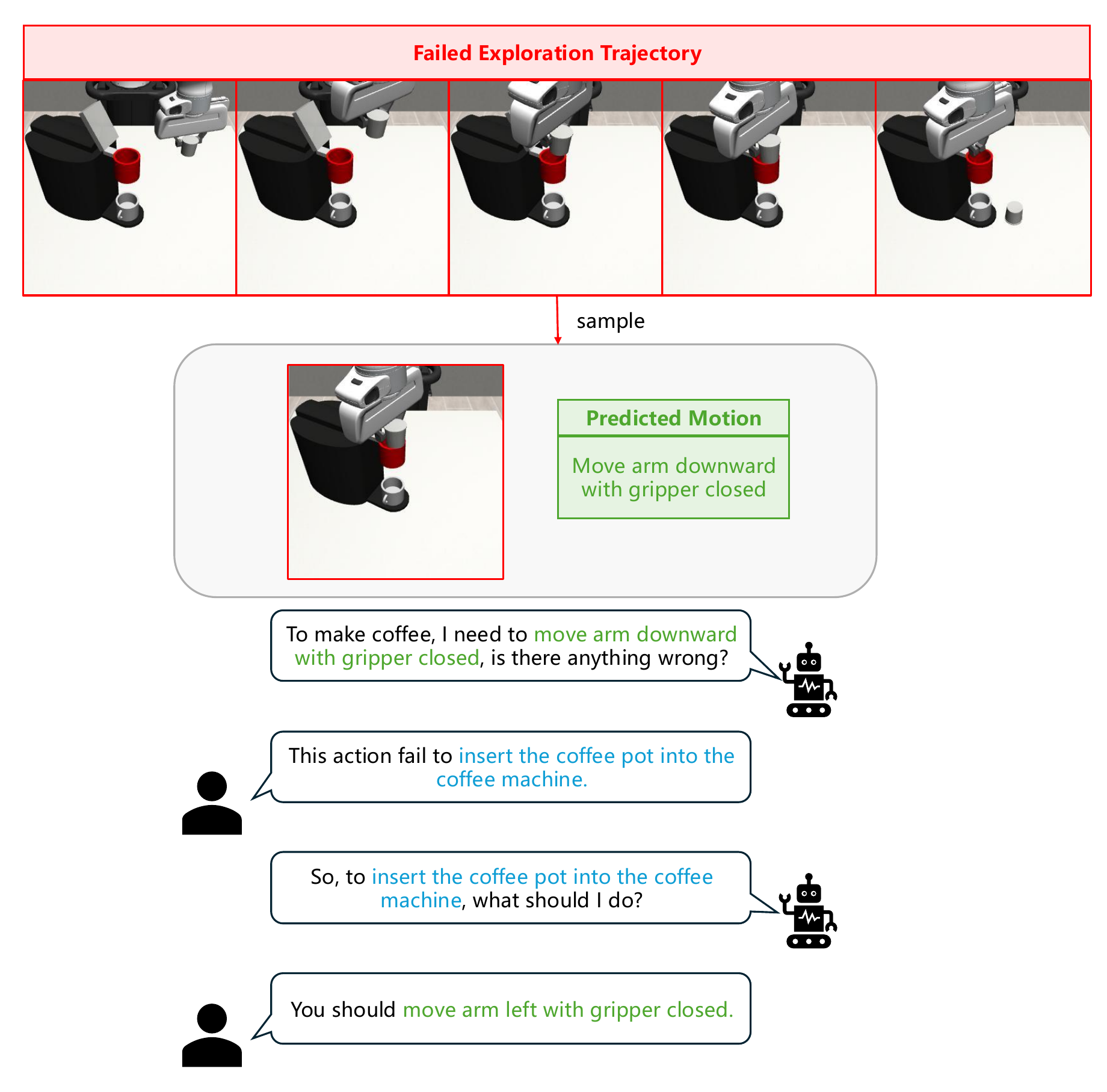}
    
    \caption{The demonstration of offline human annotation data collection process.}
    \label{fig:offline}

\end{figure}

The offline human annotation data are collected as shown in Figure~\ref{fig:offline}. We first deploy the motion prediction module to explore the environment and record the predicted motion instruction. For these collected trajectories, we sample the trajectories every 30 timestep, offering semantic feedback and adjusting the motion instructions accordingly.

\subsection{Motion-conditioned Diffusion Policy.} 
To convert the coarse-grained motion instruction into fine-grained, high-frequency robotic action, we train a multi-task, motion-conditioned diffusion policy. We take both the image observation and robotic proprioceptive as input, the image observation shape is 84, and the robotic proprioceptive consists of end-effector position, end-effector rotation, and the gripper width.
To enhance the model's temporal perception capabilities, thereby improving its ability to predict actions that adhere to the motion instructions, we integrate historical information from past 5 time steps with a temporal attention mechanism to extract temporal information. Subsequently, the observation features endowed with temporal information are used as conditional inputs in the diffusion policy. We employ 500 expert demonstrations for each task to compose the training dataset. This dataset is then used to train the diffusion policy over 200 epochs, utilizing a learning rate of 3e-4.

The learnable motion codebook is proposed to capture the discriminative features of various motion instructions. In most cases, the motion instruction predicted by MLLMs could be directly retrieved from the dictionary to obtain the corresponding language feature. However, when the predicted motion instruction is not in the dictionary, we utilize a clip text encoder to calculate the similarities between the predicted motion instruction and motion instructions in the dictionary, selecting the closest motion instruction to obtain the index.

\subsection{Evaluation Tasks}

\begin{figure*}[t]
    \centering
    \includegraphics[width=0.7\linewidth]{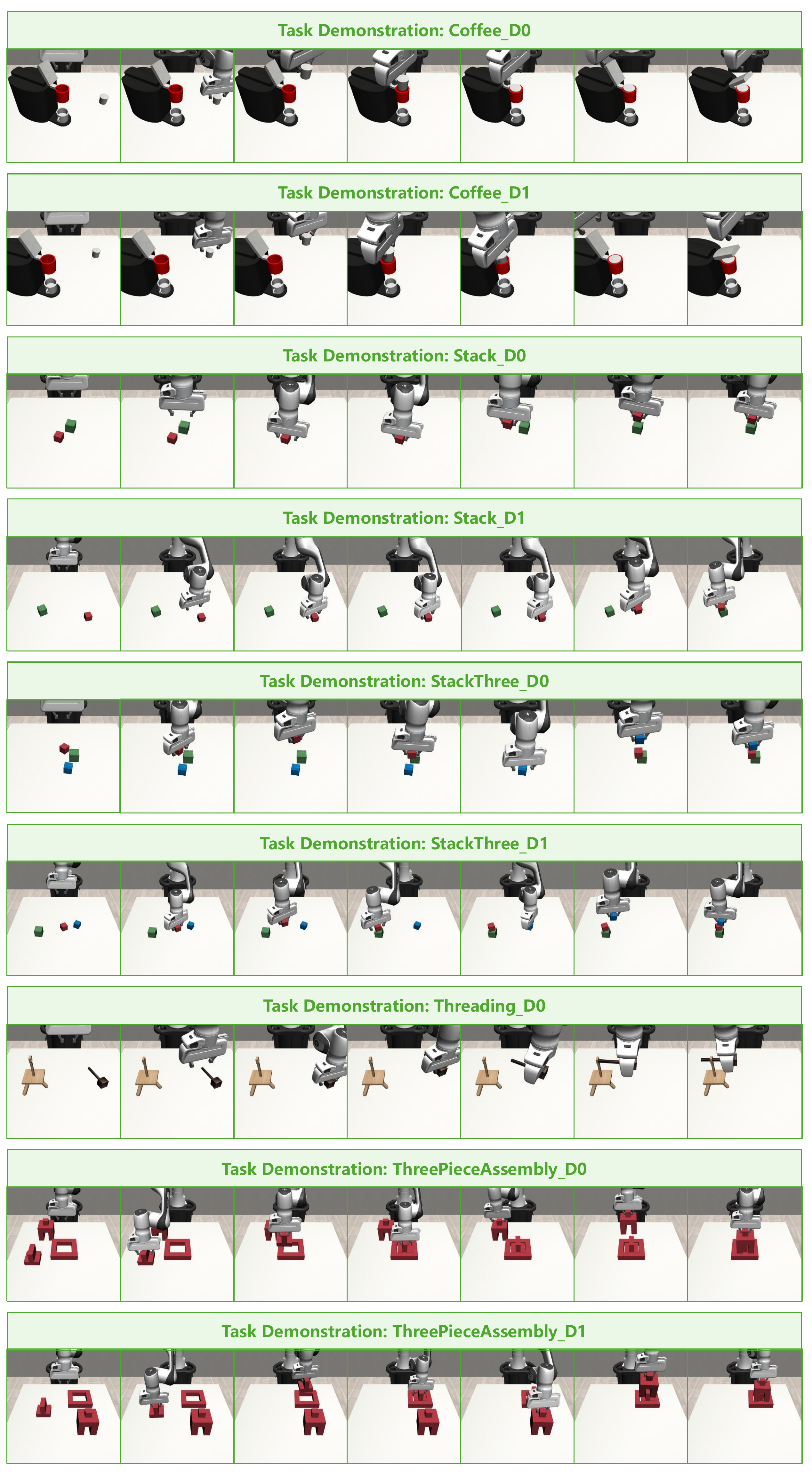}
    
    \caption{The motion instruction dataset annotation. }
    \label{fig:task}

\end{figure*}

We demonstrate 9 simulation manipulation tasks in Figure~\ref{fig:task}, which include long-horizon manipulation tasks such as ``Coffee'' and ``ThreePieceAssembly'', and fine-grained manipulation tasks such as ``Threading''. By evaluating our framework on these tasks, we could verify the effectiveness of our method.

\section{Ablation Results of Motion Codebook}

\begin{table}[t]
\centering

        \begin{tabular}{ccc}
            \toprule
            Motion Correction & Codebook  & SR \\
            \midrule
            $\times$ & $\times$ &  44.4\% \\
            $\times$ & $\checkmark$ & 46.9\% \\
            $\checkmark$ & $\times$ & 48.2\% \\
            $\checkmark$ & $\checkmark$ & 57.8\% \\
            \bottomrule
        \end{tabular}
        \caption{Ablation result of codebook}
        \label{tab:codebook}
\end{table}

\begin{figure*}[t]
    \centering
    \includegraphics[width=\linewidth]{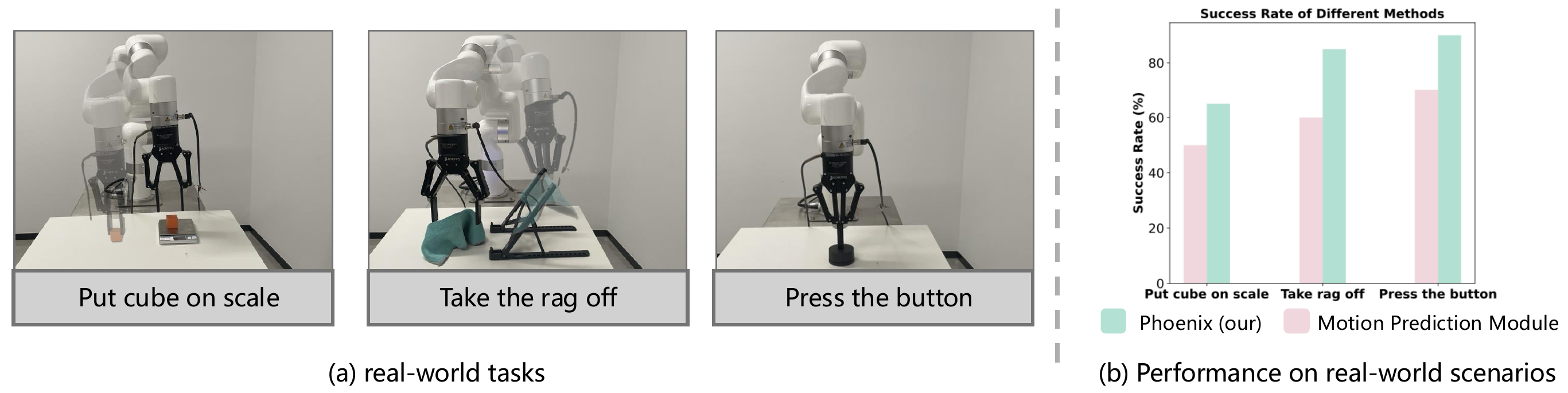}
    
    \caption{The real-world experiments. The results prove the generalization ability of our framework in real-world scenarios.}
    \label{fig:real_world}
    \vspace{-1em}
\end{figure*}

In this work, we train a motion codebook to provide discriminative motion instruction features for motion-conditioned policy. As demonstrated in Table~\ref{tab:codebook}, the policy guided by the motion codebook can better adhere to motion instructions, thus achieving better performance in manipulation tasks (44.4\% v.s. 46.9\%). Besides, benefiting from the discriminative motion instruction feature, our model could correct its action to achieve better performance (48.2\% v.s. 57.8\%) when the motion correction module is proposed to correct motion instruction.

\begin{figure*}[t]
    \centering
    \includegraphics[width=0.6\linewidth]{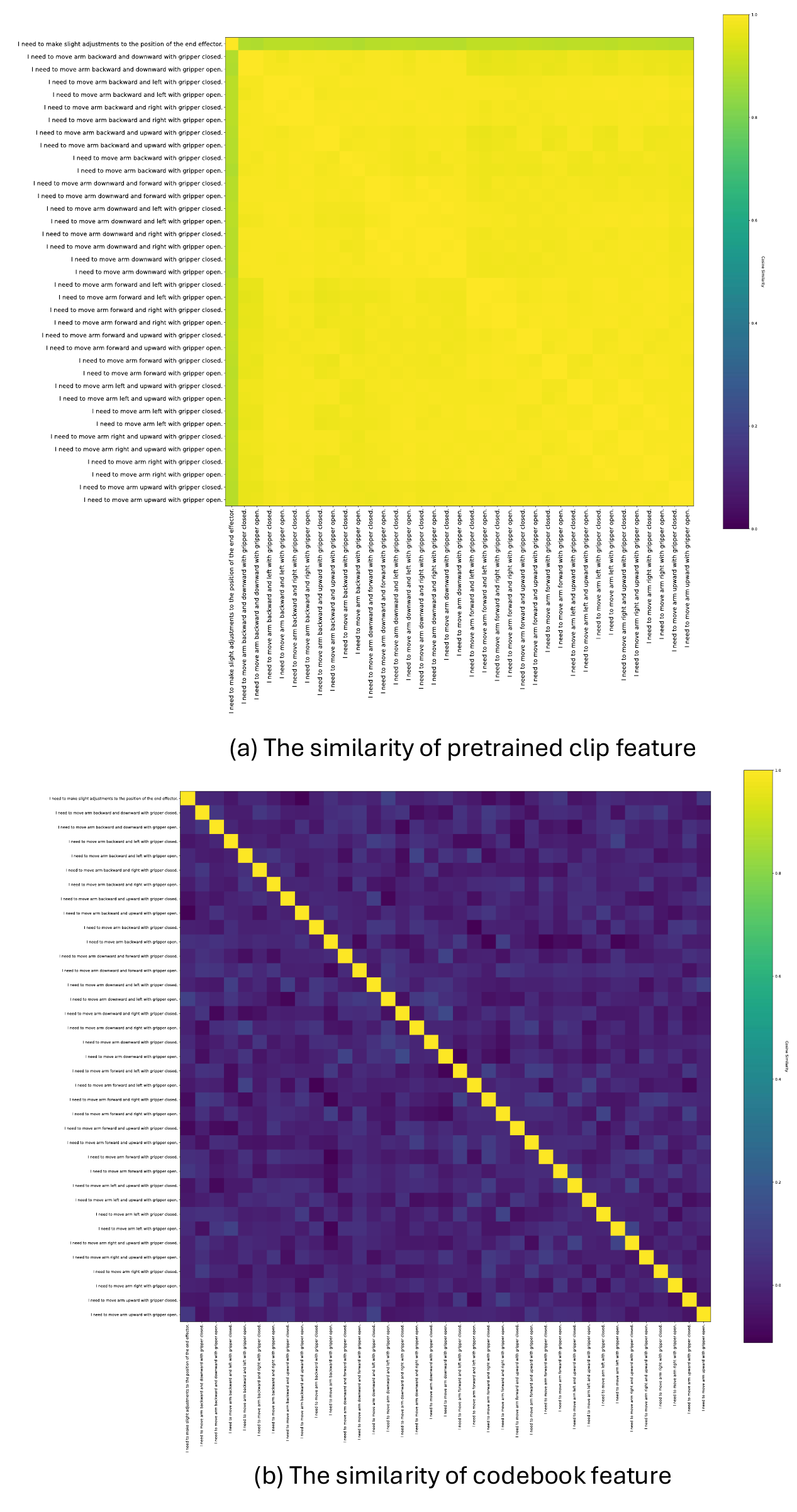}
    
    \caption{The similarity matrix of different motion instruction feature. }
    \label{fig:sim}

\end{figure*}

Furthermore, we employ the CLIP model~\cite{radford2021learning} to extract features from the motion instructions, with the resulting similarity matrix presented in Figure~\ref{fig:sim}(a). Additionally, we extract the motion instruction feature from our motion codebook, and the corresponding similarity matrix is displayed in Figure~\ref{fig:sim}(b). The similarity matrix indicates that the CLIP model struggles to effectively provide discriminative motion instruction features, as the representational similarity between any two features exceeds 90\%. In contrast, our learnable motion codebook offers discriminative representations, which facilitates the understanding of textual information for the low-level diffusion policy, thereby enhancing precise robotic action prediction.

\section{More Real-world Experiments Results}
We also prove the effectiveness of our method in rule-based manipulation policy with an xArm robot arm. As shown in Figure~\ref{fig:real_world}(a), we conduct experiments on three tasks: putting the cube on the scale, taking the rag off, and pressing the button. For each manipulation task, we collect 80 trajectories with corresponding motion instructions. To deploy the MLLMs in real-world scenarios, we fine-tuned a TinyLLaVA-OpenELM-450M-SigLIP-0.89B model~\cite{zhou2024tinyllava} to operate at a frequency of 3Hz on a 10G 4070. We also replace the diffusion policy with a rule-based operation to execute the robotic actions to adhere to the motion instructions.  During the inference process, we introduced human-in-the-loop interventions to manually correct failure situations and collect corresponding refined interaction trajectories. We collect 20 refined trajectories per task, which serve as the training dataset for the motion correction model to implement a motion-based self-reflection framework.

We conduct 20 trials and report the average success rate results in figure~\ref{fig:real_world}(b), the results prove that the motion prediction module could leverage the perceptual and inferential capabilities of MLLMs for manipulation tasks. Besides, our motion-based self-reflection model further significantly enhances the success rate with comprehensive motion adjustment, demonstrating the effectiveness of our approach in real-world scenarios.

In real-world experiments, we fine-tune a TinyLLaVA-OpenELM-450M-SigLIP-0.89B model~\cite{zhou2024tinyllava} to predict the motion instruction. We employ a third-person perspective Realsense D435 camera to acquire observational images. These images are subsequently center-cropped to shape of 384x384 and inputted into the finely-tuned TinyLLaVA model to derive motion instructions.
We collect 8 motion instructions: ``move arm upward'', ``move arm downward'', ``move arm right'', ``move arm left'', ``move arm forward'', ``move arm backward'', ``open the gripper'' and ``close the gripper''. Each movement instruction directs the arm to move 2 cm toward the target direction.

{
    \small
    \bibliographystyle{ieeenat_fullname}
    \bibliography{main}
}